\documentclass{ametsocV6.1}

\usepackage{amsmath,amssymb,amsfonts}
\usepackage{algorithmic}
\usepackage{graphicx}
\usepackage{textcomp}
\usepackage{xcolor}
\usepackage{multirow}
\usepackage{multicol}
\usepackage{amssymb}
\usepackage{array}
\usepackage{arydshln}

\newcolumntype{M}[1]{>{\centering\arraybackslash}m{#1}}

\def\BibTeX{{\rm B\kern-.05em{\sc i\kern-.025em b}\kern-.08em
    T\kern-.1667em\lower.7ex\hbox{E}\kern-.125emX}}

\title{Rainfall regression from C-band Synthetic Aperture Radar using Multi-Task Generative Adversarial Networks}

\authors{
 Aurélien COLIN, \aff{a}
 \correspondingauthor{Aurélien COLIN, acolin@groupcls.com}
 Romain HUSSON, \aff{a}
}

\affiliation{
  \aff{a}{Collecte Localisation Satellite, France, Brest}
}

\abstract{
This paper introduces a data-driven approach to estimate precipitation rates from Synthetic Aperture Radar (SAR) at a spatial resolution of 200 meters per pixel. It addresses previous challenges related to the collocation of SAR and weather radar data, specifically the misalignment in collocations and the scarcity of rainfall examples under strong wind. To tackle these challenges, the paper proposes a multi-objective formulation, introducing patch-level components and an adversarial component. It exploits the full NEXRAD archive to look for potential co-locations with Sentinel-1 data. With additional enhancements to the training procedure and the incorporation of additional inputs, the resulting model demonstrates improved accuracy in rainfall estimates and the ability to extend its performance to scenarios up to 15 m/s.
}

\begin{document}
\nolinenumbers
\maketitle

\statement{
Identifying rainfall events in coastal areas is crucial for correcting meteorological models and preventing hazards. However, quantitatively measuring precipitation in open ocean and coastal areas is challenging due to availability and range of ground-based weather radars. Satellite-based observations can provide rain information with extensive spatial coverage and high frequency but usually lack resolution. The present paper proposes to complement existing methods with Synthetic Aperture Radar (SAR). In association with observations from weather radar, it outlines an architecture to derive a relationship between SAR information and the precipitation rate.
}

\section{Introduction}

\subsection{\textcolor{black}{Context}}

Accurate rainfall estimation holds significant importance across various domains, including meteorology, hydrology, and disaster management. Rainfall can be estimated using in situ gauges, which provide continuous but localized measurements, or through remote sensing techniques. The latter commonly involves ground-based weather radar or satellite observations. While ground-based radars offer high temporal and spatial resolution, they are location-bound and impacted by topography. On the contrary, satellite observations offer global coverage with lower temporal and spatial resolution. Existing products like the {Integrated Multi-satellitE Retrievals for Gpm (}IMERG{)} \citep{10.1007/978-3-030-24568-9_19} and {Climate prediction center MORPHing method (}CMORPH{)} \citep{10.1175/1525-7541(2004)005<0487:camtpg>2.0.co;2} provide data at an approximate resolution of 10 km per pixel. Both methods combine passive low-orbit microwaves satellites with geostationary infrared sensors. Common microwaves satellites include Tropical Rainfall Measuring Mission (TRMM) \citep{10.1007/bf01029783}, the Global Precipitation Measurement (GPM) \citep{10.1002/qj.3313}, or the Defense Meteorological Satellite Program (DMSP) \citep{10.1029/94jd00961}. The low orbit satellites have a resolution of a few kilometers per pixel, but are not continuously over the zones of interest, in contrast to the geostationary satellites. Examples of geostationary instruments used for this application are the Advanced Baseline Imager of the Geostationary Operational Environmental Satellites (GOES/ABI) \citep{10.1175/bams-d-15-00230.1}, the Advanced Himawari Imager (AHI) \citep{10.1080/2150704x.2015.1066522}, or the Spinning Enhanced Visible and Infrared Imager of the Meteosat Second Generation (MSG/SEVIRI) \citep{10.1175/bams-83-7-schmetz-2}. {The Rain Rate Quantitative Precipitation Estimate (RRQPE) is available from Himawari and GOES at 2km/px \citep{10.1175/1520-0450(2001)040<1500:GMRAG>2.0.CO;2}.}

{Another example of earth-observation satellites is the Sentinel-1 constellation, part of the Copernicus program, and board Synthetic Aperture Radar (SAR) instruments \citep{10.1109/igarss.2014.6946711}. The Sentinel-1 constellation consists of several satellites in a low-polar orbit: Sentinel-1A (launched in 2014), Sentinel-1B (launched in 2016 before experiencing failure at the end of 2021), and Sentinel-1C (expected to be launched in 2024). Sentinel-1D is expected to follow at a later date. They provide observation of the ocean both for coastal areas (producing continuous slices 250 km in width, refered as “Interferometric Wideswath”, or IW) and the open ocean (producing regularly spaced observations around 20 by 20 kilometers) at a resolution of several meters per pixel, depending on the observation mode.}
SAR are able to detect heavy rain on the land by determining soil moisture \citep{10.3390/s19040802}. Over the oceans, SAR has been recognized for its ability to image the signature of hydrometeors. Their effects are numerous{, both in the air column and on the ocean surface}. 

\subsection{\textcolor{black}{Previous studies}}

Recent studies ha{ve} demonstrated the effect of atmospheric factors like the melting layer \citep{10.1016/j.rse.2020.112177}. In addition, the splash of hydrometeors generates ring waves that also increases the sea surface roughness \citep{10.1016/s0065-2687(08)60406-9}. Adversarial effects are also reported, such as a long-lasting dampening of the waves produced by heavy rainfall \citep{10.1175/1520-0477(1994)075<1183:teocsf>2.0.co;2} which decrease the sea surface roughness (SSR). {Most importantly}, rain conditions also correlate with changes in the {air circulation}, particularly under convective cells \citep{10.1016/j.rse.2016.10.015}. {These downdrafts, upon reaching the surface, lead to downbursts, which are localized strong winds that can reach to devastating speeds. They are also referred to as Straight Line Winds (SLW), as opposed to tornadic winds, and are know to increase with climate change \citep{10.1038/s41558-023-01852-9}.
The combined possibilities to estimate heavy rain patches, together with the wind maps derived from SAR at high-resolution will help quantify Straight Line Winds (SLW) and heavy rains, which is a unique contribution from SAR.}
Disentangling the intricate interactions between SSR and meteorological phenomena, particularly wind, poses challenges \citep{10.3389/fenvs.2022.1034045} and is a subject of ongoing research \citep{10.1029/2022gl102317}. 

Recent efforts have aimed to estimate rainfall from SAR observations. \citet{10.48550/ARXIV.2207.07333} segment precipitation into three levels (roughly equivalent to 1, 3 and 10 mm/h) \textcolor{black}{a fully-convolutional deep learning model trained} using {53 IW collocated} with  NEXRAD weather radar.  {However, it suffers from poor performances at wind speeds higher than 8 m/s, especially overestimation of the rain under downslope winds.} \citet{10.1109/jstars.2022.3224438} categorize SAR {patches of 5 km by 5 km from 125 IW} into {four} classes ({based on three thresholds at} 2.5, 8 and 16 mm/h) based on GPM collocations {but does not study the impact of the wind speed on the rain detection}. {Both studies opt to estimate rain intensity grades, which illustrates the difficulty in retrieving a continuous precipitation rate.} \citet{10.1109/JSTARS.2023.3255922} develop a continuous function to derive precipitation rates from {7} dual-polarization {IW of cyclones}, validated against measurements from TRMM{, which has a resolution of 0.25° x 0.25° (approximately 28 km x 28 km at the equator) and 3h. \textcolor{black}{This precipitation rate estimation} requires ancillary information of the surface currents{, winds} and waves to isolate the patterns caused by the precipitation, whose mis-estimation would hinder the rainfall estimation. In particular, the wind speed information relies on the cross-polarization radiometric signal and is therefore particularly sensitive to the estimation of the Noise Equivalent Sigma Zero (NESZ) whose azimuthal contribution is not taken into account on most of the dataset. }

These studies highlight the difficulty of obtaining {large quantity of} collocated rain information{, especially at high resolution}. However, the prospect of achieving a SAR-based rainfall product may be of interest in complementing {satellite observations} used in data assimilation frameworks{, where observations are used to re-calibrate physic models and are commonly used in hydrology \citep{10.5772/31246}. Rainfall information is notably of interest for improving early flood warnings \citep{10.3390/rs13040682, 10.1002/met.2079}, and the high resolution of SAR observations could provide valuable information on the morphology and intensity of rain cells}. \textcolor{black}{The objective of this paper is to improve the quantitative estimation of precipitation rates, particularly in challenging conditions involving strong winds and complex interactions with sea surface roughness.}

\subsection{\textcolor{black}{Challenges and strategies}}

Collocating two satellite-based sensors globally involves a time lag and deals with the low spatial resolution of space-based rain data. This issue is mitigated with ground-based observations from weather radar, though it is to the detriment of the variety of locations as only coastal regions are available for studying co-observations with SSR. 
Examples of weather radar instruments are the Next-Generation Radar (NEXRAD) \citep{heiss1990nexrad} {in the United States of America} or the {Operational Program on the Exchange of Weather Radar information (OPERA) in Europe} \citep{10.3390/atmos10060320}. This paper aims to address data scarcity by systematically collocating NEXRAD and Sentinel-1 observations. {Particular attention is given to the co-occurrence of strong wind and precipitation due to both of their extrema being infrequent \citep{10.1175/JAM2165.1, 10.1016/j.enconman.2010.06.015}, yet tremendously important for risk management, for understanding the meteorological situation, and because of their concurrent effect on SAR measurements \citep{10.1201/9781351235822-20}.}

In {e}arth {o}bservation, sensor collocation can lead to misalignment issues \citep{faiza2012review}. These misalignments often arise from discrepancies between proxy measurements and actual groundtruths. When matching NEXRAD data with ground gauges, \citep{10.1080/01431169008955114} noticed a horizontal drift ranging from 2 to 5 km due to hydrometeor advections. Similarly, while collocating lightning flashes with precipitation rates \citep{10.1175/jamc-d-12-040.1} observed misalignment, on a 5 km/px grid, between areas of active lightning and of heavy rainfall. For SAR data, the adversarial effects of precipitation on backscattering make it difficult to correct such misalignment, which can span from several hundred meters to a few kilometers. Unlike \citet{10.48550/ARXIV.2207.07333}, the present paper avoids manual realignment, which is labor-intensive and reduces the size of the data set. Instead, the paper optimizes \textcolor{black}{a deep learning model with} a multi-objective loss function to minimize pixel-level regression errors while preserving patch-level statistics. Multi-objective learning \citep{10.48550/ARXIV.1706.05098} serves as a form of regularization \citep{10.1109/igarss.2019.8898071} and ensures that the model meets predetermined properties \citep{DBLP:journals/corr/abs-2006-03653} of a known distribution. Among the additional terms of the loss, we introduce a{n adversary} loss \citep{Schonfeld_2020_CVPR} to further constrain the output distribution.

In the subsequent sections, we describe the dataset of collocated Sentinel-1 and NEXRAD co-observations. We then elaborate on the training methodology and finally present the model's evaluation and its performance analysis concerning various meteorological characteristics.

\section{Data}

The main dataset comprises 29,369 dual-polarization Interferometric Wideswath (IW) observations from the Sentinel-1 mission acquired between July 27, 2015, and April 29, 2023. IW observations typically have dimensions of 250 km width and around 180 km height. These observations include both radiometric and ancillary information, covering the following aspects:

\begin{itemize}
    \item Sea Surface Roughness (SSR), obtained from the Ground Range Detected (GRD) Level-1 products of Sentinel-1 mission. Both the co- and the cross-polarization are used due to their distinct variations in meteorological processes, such as wind dynamics \citep{10.1029/2019JC015056} or the presence of a melting layer \citep{10.1016/j.rse.2020.112177}.
    \item Noise Equivalent Sigma Zero (NESZ) of the cross-polarization channel. Since cross-polarization intensity is low in weak wind conditions, the contamination by noise reduces the relevance of the features it contains. The NESZ, which is not uniform across the observation, serves to indicate areas with lower confidence in the cross-polarization channel.
    \item Prediction of the 10-m wind speed from ECMWF's forecast model, as wind is a major factor influencing radiometric intensity. Using forecast wind speed instead of reanalysis is motivated by the necessity to run the regressor model in near-real time.
    \item Land mask, as high intensity in the radiometric channels can be caused by heavy rainfall, strong winds, or the presence of land. Ignoring the land mask would implicitly force the model to differentiate rainfall from land, and is an unnecessary constraint.
\end{itemize}

{To calculate SSR, we divided} the radar backscatters $\sigma_0^{vv}$ and $\sigma_0^{vh}$ by a geophysical model function (GMF) {as shown in Equation \ref{eq:ssr}}, assuming a wind speed of 10 m/s at a 45° angle relative to the direction of the satellite and neutral stratification. This normalization is applied to mitigate the impact of the incidence angle on radiometric data \citep{10.17882/56796}. We employ the GMF CMOD5.N \citep{10.1109/tgrs.2009.2027896} for $\sigma_0^{vv}$ and CMOD2POL \citep{10.1175/jtech-d-13-00006.1} for $\sigma_0^{vh}$.

\begin{equation}
    \begin{aligned}
        ssr^{vv} = \frac{\sigma_0^{vv}}{CMOD5.N(10m, 45°, \theta)} \\
        ssr^{vh} = \frac{\sigma_0^{vh}}{CMOD2POL(\theta)}
    \end{aligned} \label{eq:ssr}
\end{equation}

Although the native resolution of the IW GRD product is 10 m/px, the observations are downs{amp}led to 200 m/px. This reduction is implemented to alleviate hardware requirements, encompassing aspects such as data storage, I/O operations, memory utilization, and computation time.

These observations are matched with Digital Precipitation Rate (DPR) data sourced from the NEXRAD system \citep{10.7289/V5W9574V}. The DPR is accessible at a spatial resolution of 1° in azimuth and 250 m in range. Since the sensitivity of the SAR signal depends on the interaction between hydrometeors and the ocean surface, the corresponding NEXRAD data are collected at the minimal elevation angle available. However, the non-null elevation angle and the Earth curvature result in an increasing elevation of the observation with the distance to the ground station and lead to a progressive decoherence with the ocean surface situation. NEXRAD scanning occurs at six-minute intervals. The precipitation rate estimates are projected using linear interpolation to fit the SAR grid at a resolution of 200 m/px. Additional weather radar data from OPERA \citep{10.3390/atmos10060320} are used for independent tests. These products are available as composite observations with a spatial resolution of 2 km/px and a temporal resolution of 15 minutes. They are similarly projected to the SAR grid. {The weather radar information are considered as "groundtruth", i.e. the reference to estimate from the Sentinel-1 observations.}

\section{Methods}

\subsection*{Construction of the Dataset}

Each IW observation is subdivided into 25x25-kilometer {patches}. These patches are extracted with a stride equivalent to half their width, resulting in overlapping between adjacent patches. Patches are chosen only if their center is \textcolor{black}{with in} 175 km of the nearest NEXRAD station. The extraction yields a total of 1,839,142 patches. To manage this sizable amount, a reduction technique is applied. A histogram of wind speeds is built with a bin width of 0.5 m/s. Then, the maximum number of \textcolor{black}{rainless} patches in each bin is capped to 20\% of the largest bin count. This reduction strategy reduces the patch count to 545,898. {This sampling also aims at enhancing performance at high wind speeds, as under-sampling the majority class increases the relative weight of the distribution tail \citep{10.1109/CVPR52688.2022.00677}}. The outcome of {sampling}\textcolor{black}{, presented} in Figure \ref{fig:data/wind_distribution.png}
\textcolor{black}{, indicates that wind speeds lower than 1 m/s or higher than 13 m/s are not trimmed. Furthermore, only patches between 4 and 9 m/s are trimmed at a ratio higher than 4. This operation therefore allows for a reduction in the dataset size while preserving the samples at high wind speeds, where differentiating the contributions of hydrometeors and wind is expected to be more difficult and thus requires the largest dataset.}

The geographical distribution of these patches is depicted in Figure \ref{fig:data/us_distribution.png}.  {205,868 patches (37.7\% of the total) are acquired near Hawaii, 143,025 (26.2\%) on the USA's east coast, 141,054 (25.8\%) on its west coast, 28,026 (5.1\%) near \textcolor{black}{Puerto Rico} and 27,945 (5.1\%) in the Gulf of Mexico. The over-representation of data over Hawaii can be explained by \textcolor{black}{both a better overlap of the coastal seas and a higher average wind speed, which decreases the probability of being trimmed}.

\begin{figure}[]
    \centering
    \includegraphics[width=0.65\linewidth]{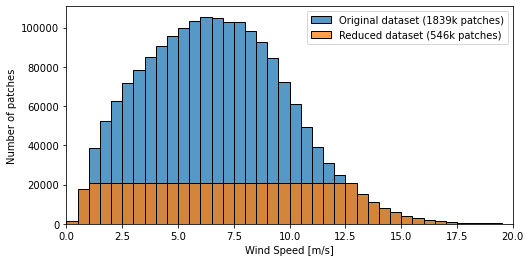}
    \caption{Wind speed distribution of the Sentinel-1/NEXRAD collocations.}
    \label{fig:data/wind_distribution.png}
\end{figure}

\begin{figure}[]
    \centering
    \includegraphics[width=0.65\linewidth]{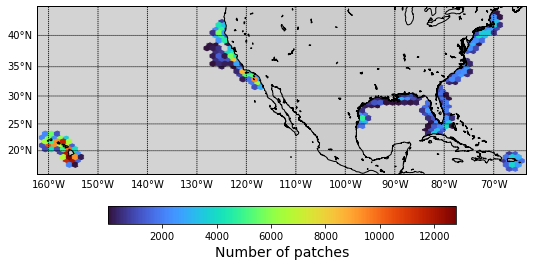}
    \caption{Geographic distribution of the patches from the Sentinel-1/NEXRAD collocations{, after the reduction operation}.}
    \label{fig:data/us_distribution.png}
\end{figure}

The patches are classified into ten distinct categories, determined by the presence of rain and the wind speed predicted by the forecast model. A patch is assigned to the "rain" category if more than 5\% of the patch area registers a precipitation rate exceeding 3 mm/h. 
\textcolor{black}{This threshold was chosen based on previous experiments which indicated that lower rainfall does not significantly correlate with higher sea surface roughness. It also aligns with a threshold used by the segmenter in \cite{10.48550/ARXIV.2207.07333}, facilitating comparison with previous models.}
Wind speed classes are established with thresholds set at 2 m/s, 6 m/s, 10 m/s, and 15 m/s. The dataset is partitioned into training (70\% of patches), validation (10\%), and test (20\%) subsets using the methodology outlined in \citet{10.48550/ARXIV.2303.09200}.
{It consists of dividing the datasets into ten subclasses (here, two precipitation conditions multiplied by five wind speed intervals) and computing their distribution for each IW. Then, we search for an IW partitioning that minimizes the difference in the 10-class distributions of the three subsets.}
This approach serves two purposes: (1) to ensure that all patches originating from a particular Sentinel-1 Level-1 {IW} product remain within the same subset and (2) to preserve the overall distribution within each subset. The former ensures that no data leak occurs between the {training, validation and test} subsets, which would risk overfitting and reduce the generalization capacity. The latter aims to reduce the risk of bias toward a particular meteorological situation. Specific quantities of patches corresponding to wind speed intervals, rain conditions, and subsets can be found in Table \ref{tab:dataset_distribution}.
\textcolor{black}{Only wind speed and precipitation conditions are included in the partitioning criteria. Adding seasonality is not favored, as it would increase the number of classes to balance to 40 instead of 10, which would become difficult if not impossible with the available number of co-locations. Moreover, seasonality mainly affects wind speed and precipitation, which are already explicitly balanced.}

\begin{table}
    \centering
    \resizebox{0.65\columnwidth}{!}{
        \begin{tabular}{|c|c|c:c:c|c:c:c|}
            \cline{3-8}
             \multicolumn{2}{c|}{} & \multicolumn{6}{c|}{\textbf{Patches with a percentage of surface with}} \\
             \multicolumn{2}{c|}{} & \multicolumn{6}{c|}{\textbf{precipitation higher than 3 mm/h...}} \\
            \cline{3-8}
             \multicolumn{2}{c|}{}  &  \multicolumn{3}{c|}{... higher than 5\%}  &  \multicolumn{3}{c|}{... lower than 5\%}  \\
            \hline
            \multirow{10}{*}{\rotatebox{90}{\parbox{2.5cm}{\centering \textbf{Maximum ECMWF wind speed [m/s]}\\[1ex]}}} & \multirow{2}{*}{[0, 2]} & 97 & 44& 76& 16,265 & 3,021& 5,536\\
             & & 45\% & 20\%& 35\%& 66\%& 12\%& 22\%\\
            \cline{2-8}
             & \multirow{2}{*}{[2, 6]} & 4,054 & 890& 1,299& 103,068& 6,300& 28,718\\
             & & 65\%& 14\%& 21\%& 70\%& 11\%& 19\%\\
            \cline{2-8}
             & \multirow{2}{*}{[6, 10]} & 6,971 &  875 & 1,808& 114,920& 19,083& 29,529\\
             & & 72\%& 9\%& 19\%& 70\%& 12\%& 18\%\\
            \cline{2-8}
             & \multirow{2}{*}{[10, 15]} & 3,899 & 520 & 1,159 & 113,707 & 21,757 & 32,648\\
             & & 70\%& 9\%& 21\%& 67\%& 13\%& 19\%\\
            \cline{2-8}
             & \multirow{2}{*}{$>$ 15} & 715 & 185& 221& 12,231& 2,505& 3,797\\
             & & 64\%& 16\%& 20\%& 66\%& 14\%& 20\%\\
            \hline
             \multicolumn{2}{c|}{} & Train & Val. & Test & Train & Val. & Test \\
            \cline{3-8}
        \end{tabular}
    }
    \caption{Number of patches from each wind speed interval, rain condition, and subset. The second number of each cell corresponds to the percentage of patches in the training, validation, and test subsets.}
    \label{tab:dataset_distribution}
\end{table}

\subsection*{Model Architecture}

The architecture of the convolutional model is depicted in Figure \ref{fig:model_architecture}. It draws inspiration from the U-Net architecture, a commonly used model for segmentation. The U-Net's fully convolutional nature ensures translation equivariance, which is vital for dividing a Sentinel-1 product into overlapping patches across the whole swath and segment the tiles. This property minimizes discontinuities between adjacent patches when reconstructing the complete observation.

The model employs both image and scalar inputs. The image input includes the SSR for both polarizations and the land mask, each at a 100 m/px resolution. The scalar input comprises the incidence angle, cross-polarization NESZ, and the a priori wind speed of the forecast model. Each scalar input provides a single value per channel per patch. Within the model, these inputs are spatialized and concatenated with activation maps after the encoder processes the image input.

In contrast to the conventional UNet model, the precipitation rate predictor generates three outputs:

\begin{itemize}
    \item The precipitation rate regression output $y_{rr}$.
    \item The binary segmentation output $y_{seg}$, identifying regions with a precipitation rate exceeding 3 mm/h.
    \item The discriminator output $y_D$, exclusively employed during training for an additional {adversary} loss.
\end{itemize}

\begin{figure}
    \centering
    \includegraphics[width=0.65\linewidth]{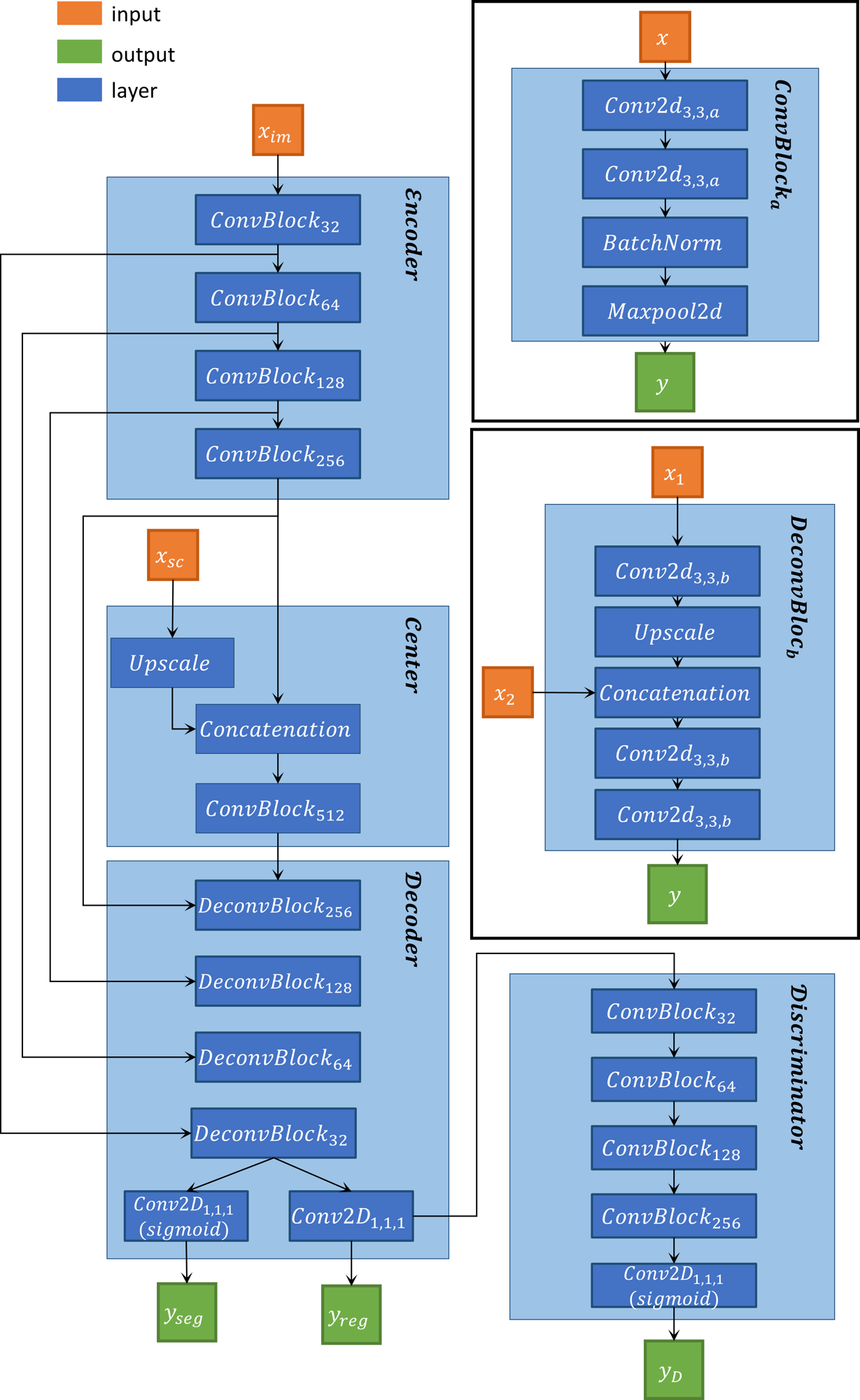}
    \caption{Architecture of the multi-objective model. Inputs encompass a 2D image $x_{im}$ and scalar $x_{sc}$ data. Outputs consist of binary segmentation $y_{seg}$, regression $y_{rr}$, and discriminator output $y_D$. Convolutional layers utilize ReLU activation functions by default. ConvBlock and DeconvBlock are presented in expanded form on the side, where the parameters $a$ and $b$ correspond to the number of convolution kernels in each convolution layer.}
    \label{fig:model_architecture}
\end{figure}

\subsection*{Training Process}

\subsubsection*{Random Sampling with Replacement from Wind/Rain Classes}

The optimization of deep learning models using gradient descent typically involves partitioning the training set into batches and iteratively updating model weights on each batch. This procedure, known as an "epoch," is repeated multiple times \citep{carney1998epoch}. In most cases, this constitutes sampling without replacement. {At the end of each epoch, performance metrics are computed in the validation set to ensure it does not diverge from those computed on the training set, which would indicate overfitting.} However, this approach treats each element of the training set uniformly, which is problematic, as certain low-probability phenomena, notably the simultaneous occurrence of rainfall and high wind speed, are crucial. As indicated in Table \ref{tab:dataset_distribution}, this meteorological scenario is underrepresented in the dataset. Employing the typical epoch-based training, the model might overemphasize processing rainless, mid-wind speed samples. To address this, class weights can be introduced for each wind/rain class based on their appearance probabilities, enhancing their impact on the derivative \citep{pmlr-v48-gopal16}. Nevertheless, this solution may introduce training instability due to the substantial disparity in class probabilities \citep{pmlr-v37-zhaoa15}.

We have therefore opted against epoch-based training and instead adopted random sampling with replacement \citep{10.48550/ARXIV.1405.3080}. Given a batch size of $b$ and $c$ representing the number of classes, each weight update batch is constructed by selecting $\frac{b}{c}$ samples from each class. The validation step, usually performed at the end of every epoch, is carried out after a specific number of weight updates. In our case, each batch contains two samples from each of the ten wind/rain classes. Validation is carried out after 512 batches, which involves 256 similarly designed random samples with replacement. The training process concludes after the 100th validation.

The discriminator is trained with the same random sampling with replacement, but only on samples closer than 80 km. The reason is the decrease of the azimuth resolution along the range, as well as the elevation of the observation that decrease the coherence of the rain signatures observed from both sensors.

\subsubsection*{Losses}

As realignment has not been performed due to the vast dataset size, the learning scheme can be considered as weakly supervised framework due to groundtruth noise. One strategy to improve model performance under weak supervision is to engage in multitask optimization where secondary tasks serve as regularizers. Consequently, the model aims to minimize the loss $\mathcal{L}$ defined in Equation \ref{eq:loss}.

\begin{align}
    \mathcal{L}(y, \hat{y}) &
        = a \cdot \mathcal{L}_{rr}(y_{rr}, \hat{y}_{rr}) + b \cdot \mathcal{L}_{seg}(y_{seg}, \hat{y}_{seg}) \nonumber \\&
        \quad + c \cdot \mathcal{L}_{max}(y_{rr}, \hat{y}_{rr}) + d \cdot \mathcal{L}_{mean}(y_{rr}, \hat{y}_{rr}) \nonumber  \\&
        \quad + e \cdot \mathcal{L}_D(\hat{y}_{rr})
        \label{eq:loss}
\end{align}

Let $n$ be the number of pixels in $y$, and $M$ represent the land mask, where $M_i = 0$ if the $i$-th pixel is over land and $M_i = 1$ if the pixel is over the ocean. We denote $I = \{i \in [1, n] | M_i = 1\}$.

\begin{align}
    \mathcal{L}_{seg}(y, \hat{y}) &= - \mathbb{E}\left[ \sum_{i\in I} y_i \cdot \log \hat{y}_i \right] \\
    \mathcal{L}_{rr}(y, \hat{y}) &= \mathbb{E}\left[ \sum_{i\in I} (y_i - \hat{y}_i)^2 \right]^\frac{1}{2} \\
    \mathcal{L}_{max}(y, \hat{y}) &= \left[max(M \cdot y) - max(M \cdot \hat{y})\right]^2 \\
    \mathcal{L}_{mean}(y, \hat{y}) &= \mathbb{E}\left[ \sum_{i\in I} (y_i - \hat{y}_i) \right] \\
    \mathcal{L}_D(\hat{y}) &= \mathbb{E}[\mathcal{D}(\hat{y})]
\end{align}

Each component of the loss serves a distinct purpose:

\begin{itemize}
    \item $\mathcal{L}_{seg}$, the segmentation loss constrains the surface area of the estimated precipitation.
    \item $\mathcal{L}_{rr}$, the precipitation rate regression loss enhances the spatial variation of precipitation rate values.
    \item $\mathcal{L}_{max}$, the maximum loss ensures accurate capture of high precipitation rate values, despite the misalignment-related fuzziness.
    \item $\mathcal{L}_{mean}$, the mean loss, drives the model to produce near-zero results for rainless pixels.
    \item $\mathcal{L}_D$, the discriminator loss, sharpens predictions.
\end{itemize}

The parameters $a$, $b$, $c$, $d$, and $e$ are set respectively to 5, $\frac{1}{15}$, $\frac{1}{40}$, $\frac{1}{40}$, and 5. GradNorm \citep{pmlr-v80-chen18a} has been tested to normalize the gradient of each loss component, but failed to enhance the performances.

The discriminator is optimized using a{n adversary} approach \citep{Schonfeld_2020_CVPR}, while both optimizations employ the RMSProp optimizer \citep{tieleman2012lecture}, which appears to be more stable on {adversary frameworks} than Adaptive Momentum (AdaM) \citep{kingma2017adam}. To avoid the numerical instabilities dues to high gradients caused by strong rainfall, we apply a gradient clipping of 1 \citep{zhang2020gradient}. The learning rate is set at $10^{-5}$.

\section{Results}

The model's performance is evaluated on two test sets. One test set corresponds to 20\% of the NEXRAD dataset placed into the test subset, maintaining a similar distribution to the training and validation subsets. A second test set is composed of data from OPERA, which combines radar products from different countries and represents a different meteorological context, covering the shores of Europe rather than the coastal areas of the United States.

{The model is evaluated from a regression point of view with the estimated precipitation rates in the first part, then considering the rain estimation as a detection problem, which enables comparison with prior models. Finally, we discuss the performance with regard to wind speed, geographical position, and the evolution of the Sentinel-1 products.}

\subsection*{Large uncertainties in precipitation rate regression}

Due to the misalignment issue, computing pixel-based metrics becomes challenging. Instead, we use the maximum rainfall estimated (either by NEXRAD or the deep learning model) on each patch of the test set.

Figure \ref{fig:data/scatter_max_rain_rate.png}a displays the distribution of the maximum rainfall reached by NEXRAD DPR (x-axis) and the Sentinel-1 regressor (y-axis) for each patch of the test set. The Pearson Correlation Coefficient (PCC), calculated as the covariance of both the estimated and the groundtruth precipitation rate divided by the product of their respective standard deviations, reaches 77\%. However, the figure shows a wide distribution with a Root Mean Square Error (RMSE) of 12.18 mm/h. The linear regression has a slope of 0.69, indicating an underestimate of high rainfall. {This underestimate is visible in Figure \ref{fig:data/scatter_max_rain_rate.png}b as the Sentinel-1 regressor appears to produce a lower proportion of rates between 30 and 100 mm/h. On the contrary, rates between 1 and 3 mm/h, as well as between 10 and 30 mm/h are over-represented in the model distribution.}

\begin{figure}[]
    \centering
    \begin{tabular}{M{0.45\linewidth}M{0.45\linewidth}}
        {(a)} & {(b)} \\
    \end{tabular}
    
    \includegraphics[width=0.95\linewidth]{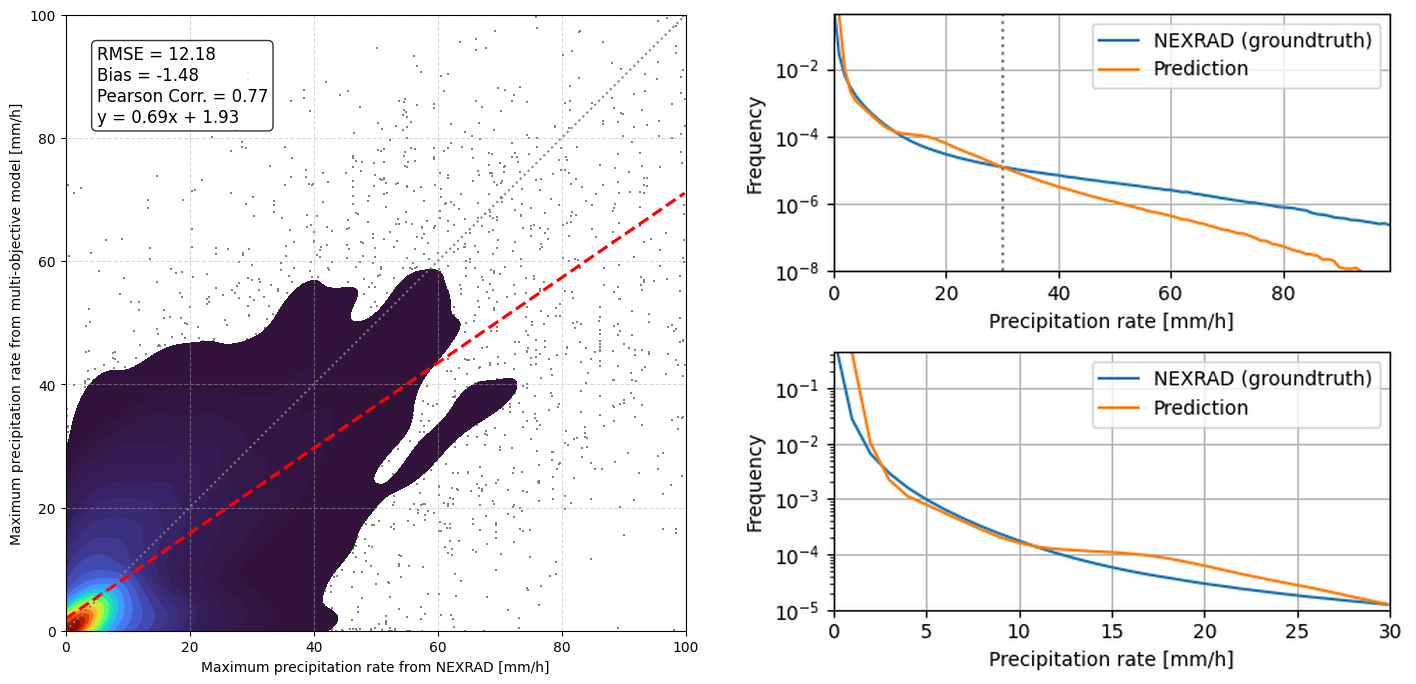}
    
    \caption{{(a)} Distribution of the maximum NEXRAD DPR rainfall on the patches of the test set (x-axis) against the maximum predicted rainfall (y-axis). The colored area accounts for 90\% of the rain patches, while the remaining 10\% are indicated as points. {(b) Pixel-wise distribution of the precipitation rate from the NEXRAD dataset and the model prediction between 0 and 100 mm/h (top) and zoomed in on 0 to 30 mm/h (bottom).}}
    \label{fig:data/scatter_max_rain_rate.png}
\end{figure}

\begin{figure*}[]
    \centering
    \includegraphics[width=0.9\linewidth]{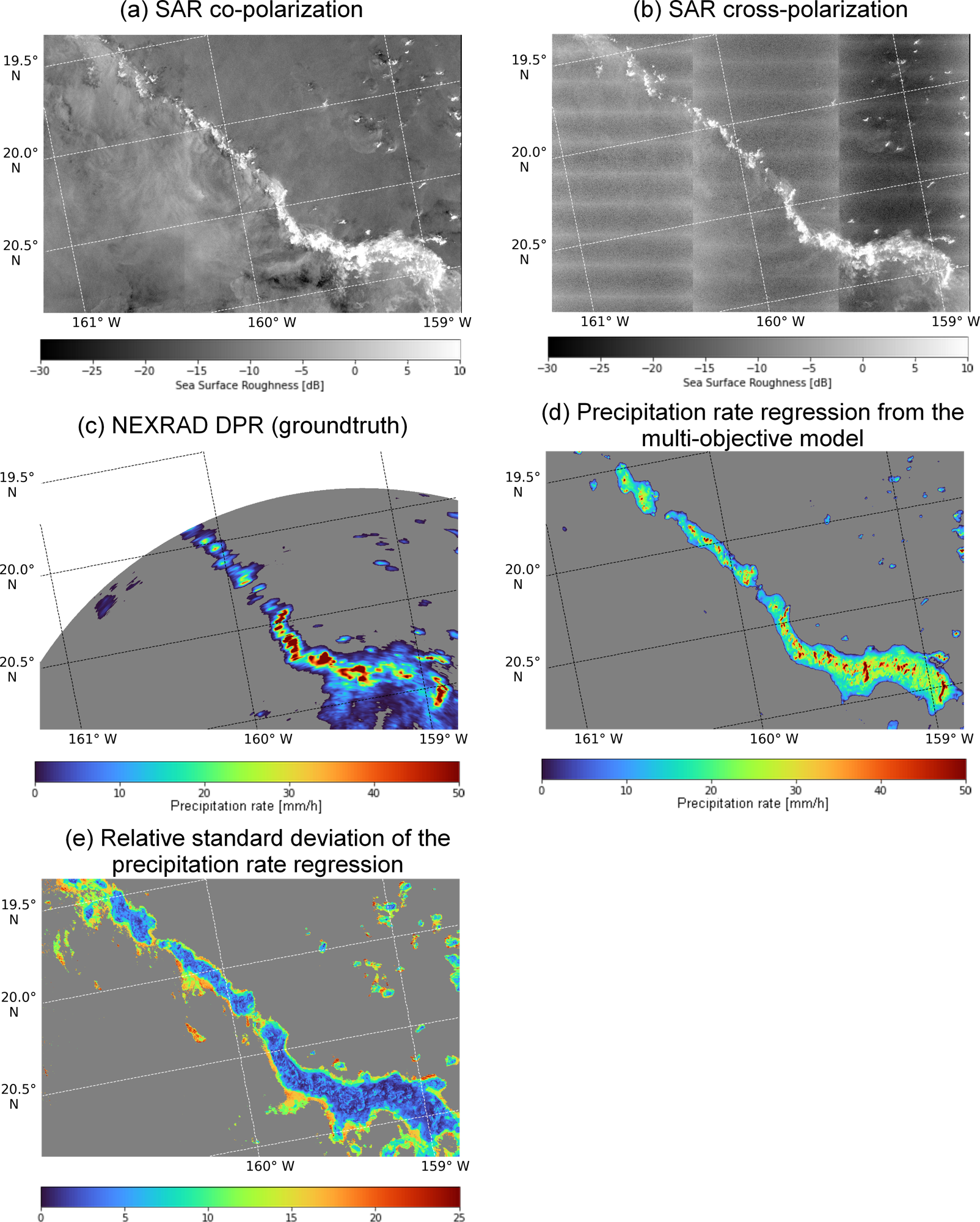}
    \caption{Co-observation from both Sentinel-1 (acquired on 2018-02-05 at 04:47:13 UTC) and NEXRAD (acquired at 04:48 UTC).}
    \label{fig:20180205t044713}
\end{figure*}

{The over-representation of rates around 20 mm/h is also} visible in {the case study of} Figure \ref{fig:20180205t044713}. Overall, the rain signatures that appear on the SAR observations are correctly delimited. The high-precipitation areas of the squall lines are also estimated at precipitation rates consistent with the weather radar. 
{However, the distribution of the precipitation rate differs between the estimation from the regression model and the NEXRAD observation. It plateaus at around 25 mm/h before quickly reducing to zero. The overall extent of the rain area is therefore smaller in the prediction. These differences indicate shortcomings of the adversarial loss, which penalizes the model when the distribution of the groundtruth and the regressor differ.}

{To evaluate the uncertainty of the model, and reduce the impact of stochastic steps of the model (such as the initialization and the order of the samples), the evaluation i\textcolor{black}{s} performed on five iterations of the training process. The mean and standard deviation across the trainings are computed for each pixels. The relative standard deviation is then calculated as the ratio of the standard deviation and the mean prediction. In (e), it indicates confidence in the rainfall prediction, though the delineation of the border of the squall line can be fuzzy.}

\begin{figure*}[]
    \centering
    \includegraphics[width=0.9\linewidth]{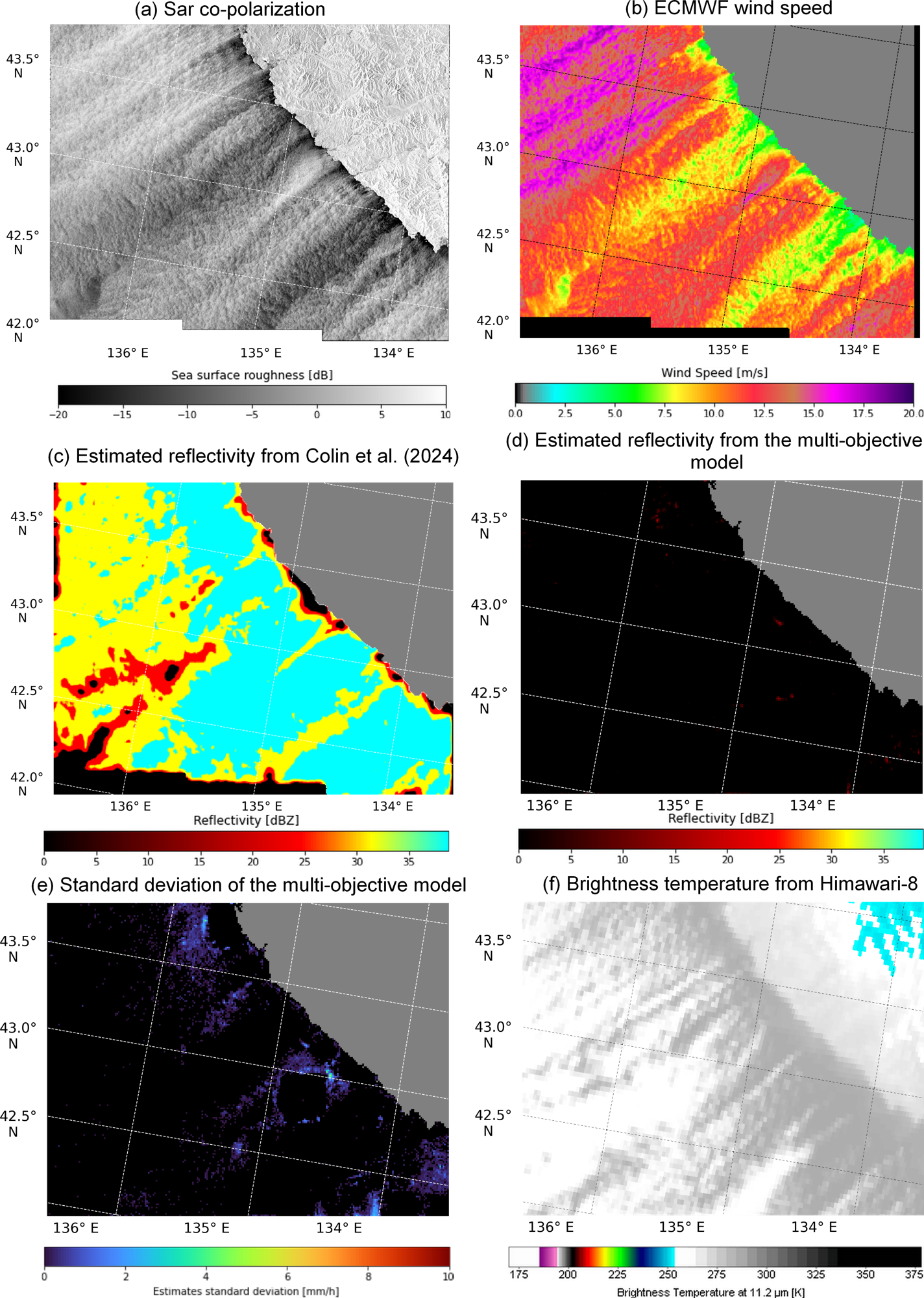}
    \caption{Observation from Sentinel-1 (acquired on 2021-10-16 at 21:05:33 UTC). The precipitation rate from the multi-objective model is converted in the equivalent reflectivity provided in \citet{10.48550/ARXIV.2207.07333}.}
    \label{fig:20211016T210533}
\end{figure*}

In Figure \ref{fig:20211016T210533}, the multi-objective model is compared to the four-level segmentation obtained from \citep{10.48550/ARXIV.2207.07333}. In this scene, the katabatic wind generates complex patterns on the ocean surface {(\textcolor{black}{Figure \ref{fig:20211016T210533}}a)}, recognized as heavy rainfall by the {existing segmenter} {(\textcolor{black}{Figure \ref{fig:20211016T210533}}c)}. {In \textcolor{black}{Figure \ref{fig:20211016T210533}}f, the brightness temperature from Himawari-8's band at 11.2 µm, from which precipitation rates can be derived \citep{10.1175/bams-d-15-00230.1}, does not indicate any rainfall event. In particular, the derived Rain Rate Quantitative Precipitation Estimate (RRQPEF) is uniform at 0 mm/h over the whole observation. The overestimate from the segmenter} was caused by the low amount of samples with strong wind in the training set, whose size was limited by the need for manual realignment. As this condition was relaxed with the multi-objective model, the training set encompasses a more diverse wind distribution. It is therefore able to avoid the overestimate {(\textcolor{black}{Figure \ref{fig:20211016T210533}}d)}. The standard deviation {(\textcolor{black}{Figure \ref{fig:20211016T210533}}e)} indicates stability across various random initializations.

\subsection*{High performances in rainfall detection}

Under a categorization framework, the patches are categorized as containing rain based on {a} threshold of 3 mm/h on more than 5\% of the patch area. To estimate the capacity of the rain detector, we use the $F_1$ score. The $F_1$ score is the harmonic mean of precision {(i.e. success ratio, Equation \ref{eq:precision})} and recall {(i.e. probability of detection, Equation \ref{eq:recall})}. It takes into account both over- and under-detection and is less sensitive to data imbalance than accuracy. 

Given a categorization array $z$ of $n$ elements, where each element $z_i$ equals 1 if patch $i$ contains rainfall, 0 if patch $i$ is rainless, and $\hat{z}$ the corresponding prediction, the formulation of the $F_1$ score is given in Equation \ref{eq:f1}.

\begin{equation}
    precision(z,\hat{z}) = \frac{\sum\limits_{i=1..n} z \cdot \hat{z}}{\sum\limits_{i=1..n} \hat{z}} \label{eq:precision}	
\end{equation}

\begin{equation}
    recall(z, \hat{z}) = \frac{\sum\limits_{i=1..n} z \cdot \hat{z}}{\sum\limits_{i=1..n} z} \label{eq:recall}	
\end{equation}

\begin{equation}
    F_1(z, \hat{z}) = 2\cdot\frac{precision(z, \hat{z})\cdot recall(z, \hat{z})}{precision(z, \hat{z}) + recall(z, \hat{z})}	
    \label{eq:f1}	
\end{equation}

\subsubsection*{Relative Importance of the Inputs}

To study the importance of each input, we train the deep learning model with various input combinations by removing a single input channel and comparing the $F_1$ score to the full architecture. The results are presented in Table \ref{tab:f1_against_input} as the mean $F_1$ score (across five training with random initialization) and its standard deviation for different wind speed intervals and inputs. {The average in the last column is computed over the indicated wind speed intervals (rather than on the patches) to account for the differences in frequency magnitude of each wind situation, which would otherwise be marginally impacted by the performances at high wind speeds.}

This analysis reveals the {model with all inputs} reaches higher $F_1$ score on average. However, various observations can be done on specific wind intervals. Without the co-polarization input (VV), the {model} has better performances on wind speeds higher than 16 m/s. This can be explained by the saturation of the SSR due to the wind's effect, which caused the disappearance of rain signatures \citep{10.1109/tgrs.2017.2732508}. The cross-polarization channel is less susceptible to saturation and remains useful under stronger winds. Furthermore, averaging over all wind speed ranges, the removal of the VH channel appears more detrimental than the removal of the VV channel. Contrary to expectations, the NESZ channel is significantly more important at high wind speeds than for weaker winds. This can be linked to the lower relevance of the VH channel under weak wind conditions. {Therefore, the normalized radar cross-section (NRCS) in the cross-polarization (and its separation with the NESZ) is only required at higher wind speeds}. Finally, the wind speed prior appears to be detrimental at high wind speeds. {This may indicate that the error in the prior, provided by the ECMWF deterministic atmospheric model, is less reliable at high wind speeds. This model is known to underestimate strong winds \citep{10.1029/2023JD039673} and shares this behavior with ECMWF’s reanalysis \citep{10.1175/jtech-d-12-00240.1}. However, since the deep learning model is trained on the prior rather than on true wind speeds (which are inaccessible), a non-unit slope between the real wind speed and the prior should not be detrimental. The negative impact of the wind speed prior at high wind speeds may be better explained by the smaller spatial and temporal extent of the strong wind phenomena. As such, the difference between the time of the atmospheric model and that of the observation, a slight spatial drift in the position of the wind event, or the resolution difference between the atmospheric model and the SAR observation would create higher discrepancies at high wind speeds than in more common situations}. Overall, using all six inputs seems to yield the best results, though only marginally in comparison with the model without the wind speed prior. The {model with all inputs} also obtains the lowest variability across the trainings.

\begin{table}[]
    \centering
    \caption{
    $F_1$ score of variants of the multi-objective model, excluding one input channel. The last raw is the original model with all inputs. Results are given for a rainfall threshold of 3 mm/h, as mean ($\mu$) and standard deviation ($\sigma$) on five training. The highest values for each wind range are highlighted in bold. {VV (resp. VH) refers to the co- (resp. cross-) polarization channel, MASK to the ground mask, NESZ to the Noise Equivalent Sigma Zero of the VH channel, INC to the observing incidence and WSPD to ECMWF's wind speed.}}
    \label{tab:f1_against_input}
    \resizebox{0.65\columnwidth}{!}{
    \begin{tabular}{|l|cc|c|c|c|c|c|c|}
        \cline{4-8}
        \multicolumn{3}{c|}{} & \multicolumn{5}{c|}{\textbf{Wind speed interval [m/s]}} \\\cline{4-9}
        \multicolumn{3}{c|}{} & [0-4] & [4-8] & [8-12] & [12-16] & [16-20] & Average \\\hline
        \cline{2-9}
        \multirow{12}{*}[12pt]{\rotatebox{90}{\parbox{2.5cm}{\centering \textbf{Removed Input}\\[1ex]}}}             & \multirow{2}{*}[2pt]{VV}   
                & $\mu$ & 0.613 & 0.599 & \textbf{0.572} & 0.462 & 0.401 & 0.529 \\
               && $\sigma$ & 0.020 & 0.013 & 0.033 & 0.045 & 0.104 & 0.036 \\
            \cline{2-9}
            & \multirow{2}{*}[2pt]{VH}
                & $\mu$ & 0.672 & 0.565 & 0.484 & 0.367 & 0.316 & 0.481 \\
               && $\sigma$ & 0.016 & 0.134 & 0.198 & 0.177 & 0.194 & 0.132 \\
            \cline{2-9}
            & \multirow{2}{*}[2pt]{MASK}
                & $\mu$ & 0.672 & 0.574 & 0.535 & 0.473 & 0.285 & 0.508 \\
               && $\sigma$ & 0.051 & 0.052 & 0.036 & 0.017 & 0.143 & 0.043 \\
            \cline{2-9}
            & \multirow{2}{*}[2pt]{NESZ}
                & $\mu$ & \textbf{0.715} & 0.593 & 0.502 & 0.437 & 0.239 & 0.497 \\
               && $\sigma$ & 0.026 & 0.003 & 0.057 & 0.062 & 0.047 & 0.025 \\
            \cline{2-9}
            & \multirow{2}{*}[2pt]{INC}
                & $\mu$ & 0.705 & 0.597 & 0.543 & 0.467 & 0.309 & 0.524 \\
               && $\sigma$ & 0.028 & 0.038 & 0.042 & 0.036 & 0.110 & 0.030 \\
            \cline{2-9}
            & \multirow{2}{*}[2pt]{WSPD}
                & $\mu$ & 0.657 & 0.602 & 0.547 & \textbf{0.498} & \textbf{0.409} & 0.543 \\
               && $\sigma$ & 0.079 & 0.018 & 0.036 & 0.019 & 0.133 & 0.034 \\
            \hline
        \multicolumn{2}{|c}{All inputs}
            & $\mu$ & 0.689 & \textbf{0.616} & 0.565 & 0.492 & 0.356 & \textbf{0.544} \\
        \multicolumn{2}{|c}{}
           & $\sigma$ & 0.025 & 0.034 & 0.047 & 0.017 & 0.077 & 0.022\\
        \hline
    \end{tabular}}
\end{table}

\subsubsection*{Impact of the Losses}

We also compare the impact of different loss components by training models with multiple loss formulations. Similarly with the previous section, variants of the original model are trained by removing a single loss, and compared with the full architecture. The results of this analysis are presented in Table \ref{tab:f1_against_loss}. 

It appears that the most crucial component of the loss is the mean over the patch. The removal of this loss led to the lowest $F_1$ score on every wind speed intervals. This component ensures that the overall rain estimation remains low to prevent overestimation. The second most important parameter is the maximum over the patch, which, in contrast to the previous component, pushes the estimates to higher precipitation rates. We can note that these two components are the patch-based statistics used to avoid the manual re-alignment of NEXRAD and Sentinel-1 data. The regression and discriminator components seem to have a similar impact on the $F_1$ score. We observed that removing the discriminator (and thus the adversarial component) reduces the standard deviation of the architecture. The component with the least impact on performance is the segmentation loss. However, the model incorporating all the losses achieves higher $F_1$ score values and lower variability between different trainings.

\begin{table}[]
    \centering
    \caption{
    $F_1$ score of variants of the multi-objective model, excluding one loss. The last raw is the original model with all losses. Results are given for a rainfall threshold of 3 mm/h, as mean ($\mu$) and standard deviation ($\sigma$) on five training. The highest values for each wind range are highlighted in bold}
    \label{tab:f1_against_loss}
    \resizebox{0.65\columnwidth}{!}{
    \begin{tabular}{|l|cc|c|c|c|c|c|c|}
        \cline{4-8}
        \multicolumn{3}{c|}{} & \multicolumn{5}{c|}{\textbf{Wind speed interval [m/s]}} \\\cline{4-9}
        \multicolumn{3}{c|}{}  & [0-4] & [4-8] & [8-12] & [12-16] & [16-20] & Average \\\hline
        \cline{3-9}
        \multirow{12}{*}[12pt]{\rotatebox{90}{\parbox{2.5cm}{\centering \textbf{Removed loss\\component}\\[1ex]}}}
            & \multirow{2}{*}[2pt]{$\mathcal{L}_{seg}$}   
                & $\mu$ & 0.674 & 0.583 & 0.542 & 0.482 & 0.338 & 0.524 \\
               &&$\sigma$ & 0.036 & 0.034 & 0.049 & 0.064 & 0.092 & 0.042 \\
            \cline{2-9}
            & \multirow{2}{*}[2pt]{$\mathcal{L}_{rr}$}   
                & $\mu$ & 0.689 & 0.586 & 0.490 & 0.408 & 0.307 & 0.496 \\
               &&$\sigma$ & 0.028 & 0.056 & 0.131 & 0.138 & 0.131 & 0.082 \\
            \cline{2-9}
            & \multirow{2}{*}[2pt]{$\mathcal{L}_{max}$}   
                & $\mu$ & 0.640 & 0.498 & 0.391 & 0.309 & 0.202 & 0.408 \\
               &&$\sigma$ & 0.111 & 0.110 & 0.106 & 0.111 & 0.082 & 0.097 \\
            \cline{2-9}
            & \multirow{2}{*}[2pt]{$\mathcal{L}_{mean}$}   
                & $\mu$ & 0.419 & 0.445 & 0.376 & 0.278 & 0.228 & 0.349 \\
               &&$\sigma$ & 0.156 & 0.110 & 0.093 & 0.106 & 0.151 & 0.095 \\
            \cline{2-9}
            & \multirow{2}{*}[2pt]{$\mathcal{L}_D$}   
                & $\mu$ & \textbf{0.691} & 0.565 & 0.523 & 0.450 & 0.272 & 0.500 \\
               &&$\sigma$ & 0.017 & 0.035 & 0.036 & 0.030 & 0.063 & 0.018 \\
            \hline
        \multicolumn{2}{|c}{$\mathcal{L}$}
            & $\mu$ & 0.689 & \textbf{0.616} &\textbf{ 0.565} & \textbf{0.492} & \textbf{0.356} & \textbf{0.544} \\
        \multicolumn{2}{|c}{}
            &$\sigma$ & 0.025 & 0.034 & 0.047 & 0.017 & 0.077 & 0.022\\
        \hline
    \end{tabular}}
\end{table}

\begin{figure*}[]
    \centering
    \includegraphics[width=0.7\linewidth]{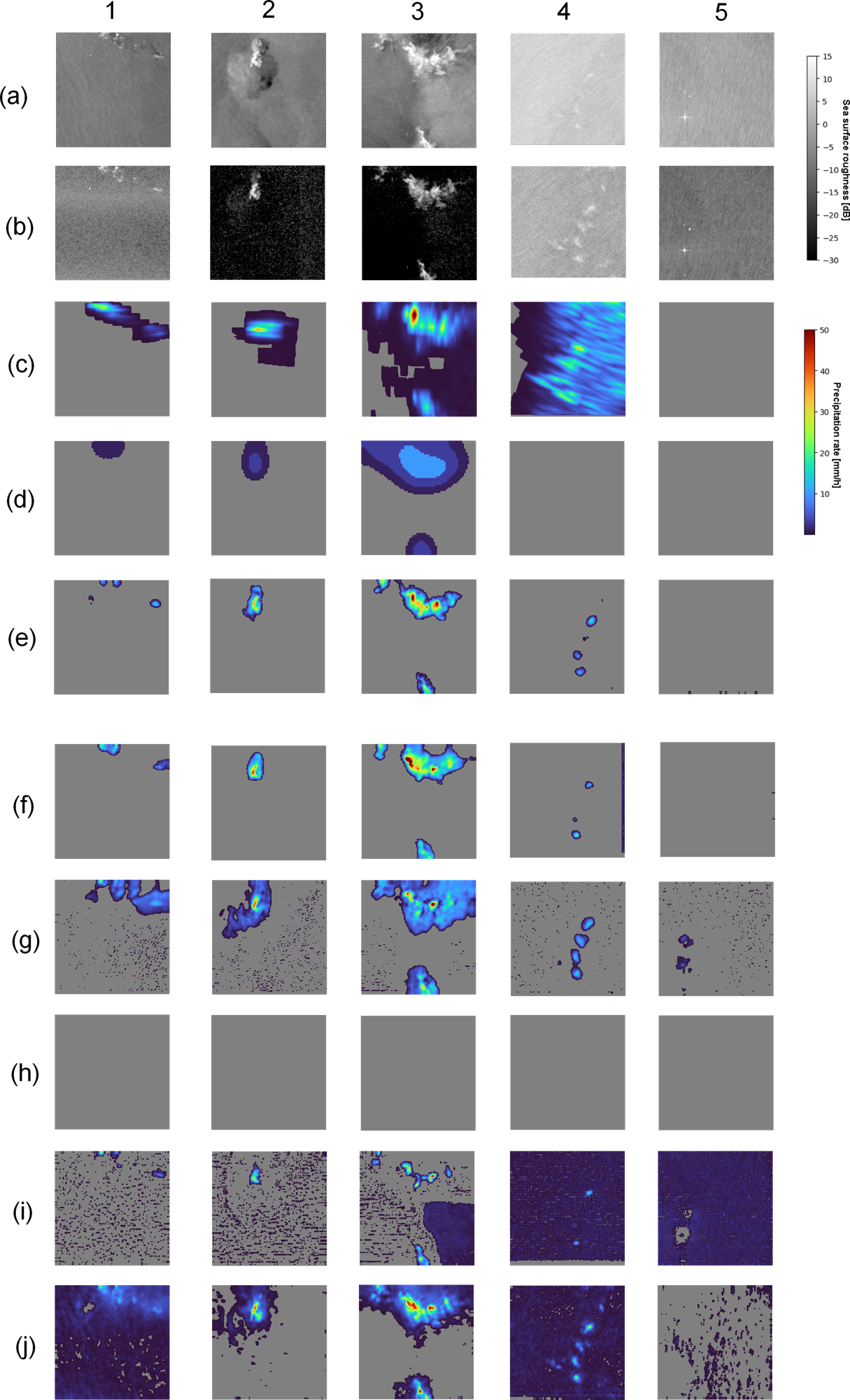} 
    \caption{Examples of rainfall regression on patches of 25x25 km. (a) the co-polarization channel in dB, (b) the cross-polarization channel in dB, (c) the groundtruth from NEXRAD, (d) the precipitation rates predicted by \citet{10.48550/ARXIV.2207.07333}, (e) by our regression model, and the models trained without $\mathcal{L}_{seg}$ (f), $\mathcal{L}_{rr}$ (g), $\mathcal{L}_{max}$ (h), $\mathcal{L}_{mean}$ (i), and  $\mathcal{L}_{D}$ (j).}
    \label{fig:thumbs_loss}
\end{figure*}

The effect of the choice of the loss is depicted at the patch level in Figure \ref{fig:thumbs_loss}. The lines (a) and (b) represent the SAR observation respectively in co- and cross-polarization. Line (c) is the precipitation rate indicated by NEXRAD's DPR. Lines (f) to (j) are the predictions by models trained by setting the weight of one loss to 0. precipitation rates lower than 0.1 mm/h are grayed out. Setting $\mathcal{L}_{max}$'s weight to zero (h) led the model to systematically return null values for precipitation rate, due to the lower frequency of rainfall among the pixels, despite the patch-level balance. Removing either $\mathcal{L}_{rr}$ (g), $\mathcal{L}_{mean}$ (i), or $\mathcal{L}_D$ (j) led to noisy segmentation with pixels overshooting the threshold. This rainfall overestimate decreases the precision metric and therefore negatively impacts the $F_1$ score. The removal of $\mathcal{L}_{{s}eg}$ produces estimates close to the {model with all losses}, which is coherent with {Table} \ref{tab:f1_against_loss}. Overall, the model trained with all losses (line (e)) generated sharper estimations than the segmenter (line (d)) and is able to provide quantitative information on the rainfall, but seems to underestimate the rainfall, especially with stratiform rainfall (column 4).

\subsection*{Performances Dependent on Meteorological Conditions}

In Figure \ref{fig:vanilla_vs_new}, the model is evaluated as a rain {classifier}, categorizing a "rain" patch as more than 3 mm/h of estimated rainfall on more than 5\% of the surface of the patch. The performances are compared to the wind speed and the {Instrument Processing Facility} (IPF) version in Figure \ref{fig:vanilla_vs_new}.  {The IPF (Instrument Processing Facility) is the software, internal of the Sentinel-1 mission, used to process Level-1 and Level-2 SAR products from the raw data received from the sensor. It is regularly updated to enhance the quality of the products, such as correcting the calibration of the sensor or the estimation of thermal noise. However, modifications in the processing of L1-GRDH products create heterogeneity for the SAR features used by the NN model., as products disseminated for the Copernicus program are not reprocessed because of the size of the required computation.} The variation of the performances with the evolution of the input distributions{, here the SAR information,} is called "data drift". It is crucial to monitor this drift to ensure service continuity. 

The overestimation at high speed, indicated by a decrease in precision ({driven by the higher} the false positive rate) at wind speeds higher than 8 m/s with the U-Net segmenter, is mitigated. The precision remains stable until 15 m/s. The underestimation on some patches, indicated by the recall is comparable to {the existing segmenter}. The overall performances, indicated by the $F_1$ score, are therefore less degraded at high wind speeds. {At wind speeds lower than 8 m/s, the $F_1$ score only marginally increases.}

We can observe that the precision shows large variations depending on the IPF version. However, it is important to note that the meteorological distribution associated with each IPF version differs as they correspond to specific timeframes. As an example, IPF 3.40, which ran from 2021-11-04 to 2022-03-23, mainly differs from IPF 3.52 by the Level-2 processing, which does not impact the present studies. The gap between the $F_1$ score of both versions, though contained within the confidence interval of IPF 3.40, is therefore mainly attributable to the different meteorological situations. On a side note, the processing may be impacted by changes in the configuration files and auxiliary files not indicated by the release of a new IPF version. Therefore, the IPF may not reflect perfectly the changes in the model inputs. Overall, the regressor shows to be resilient to changes in the IPF and is expected to be used without retraining on future versions of the IPF.

\begin{figure}[]	
    \centering	
    \includegraphics[width=0.9\linewidth]{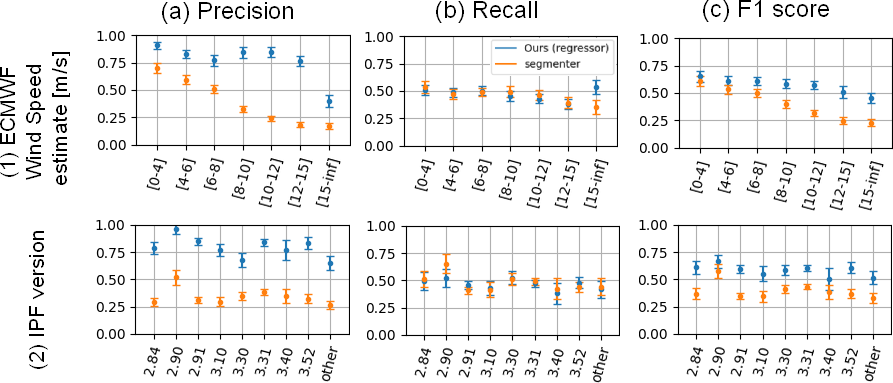}
    
    \caption{Performances of the multi-objective model (blue) compared to the segmentation model from \citet{10.48550/ARXIV.2207.07333} (orange) depending on ECMWF's forecast wind speed and the IPF version. The metrics are adjusted for each bin using likelihood weighting on the rainfall distribution \citep{10.1080/00401706.1995.10484303}. Confidence interval are given at 95\% of confidence using bootstrapping \citep{10.1214/aos/1176344552}.}	
    \label{fig:vanilla_vs_new}	
\end{figure}


In Figure \ref{fig:data/f1map.png}, the $F_1$ score is averaged for each NEXRAD station. The rain detector appears to have higher performance in the Gulf of Mexico and lower performance on the west coast of the United States. Additionally, these lower performances are linked to an higher standard deviation and a lower number of rain observations.

\begin{figure}[]
    \centering
    \includegraphics[width=0.65\linewidth]{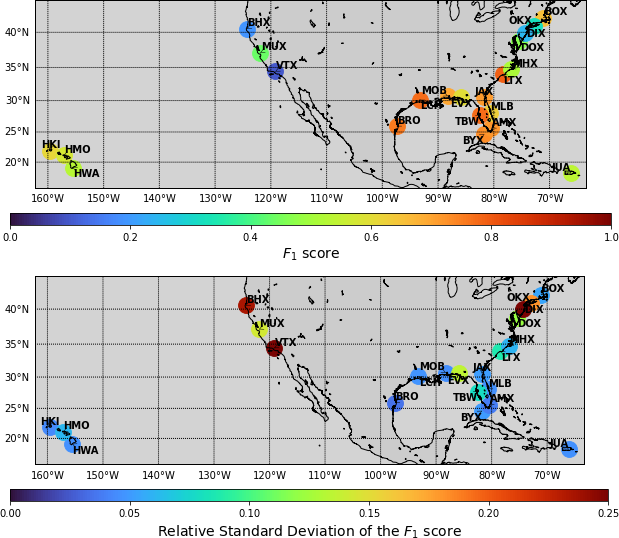}
    
    \caption{$F_1$ score of the multi-objective model for each NEXRAD ground station (top) as well as the standard deviation obtained through bootstrapping (bottom).}
    \label{fig:data/f1map.png}
\end{figure}

In Figure \ref{fig:data/rain_vs_estimated.png}, the regression and the groundtruth are accumulated for each NEXRAD station. Overall, the fraction of patches with detected rainfall matches between NEXRAD and the SAR detector. The stations BHX, MUX, and VTX, three stations of the United States' west coast, are among those with the lowest precipitation rate estimated by NEXRAD. In addition to a smaller rainfall amount, those location are characterized by different types of rain events as they are more prone to stratiform precipitation \citep{10.1029/2023gl102786}.

\begin{figure}[]
    \centering
    \includegraphics[width=0.75\linewidth]{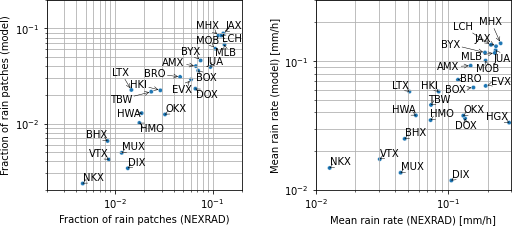}
    \caption{Left: Fraction of patches with more than 3 mm/h over more than 5\% of their surface. Right: mean rainfall. The x-axis are estimated by NEXRAD DPR and the y-axis from SAR observations by the multi-objective model.}
    \label{fig:data/rain_vs_estimated.png}
\end{figure}

A similar comparison is performed with data from the European radar project OPERA {in Figure \ref{fig:eumap}}. It indicates that the performances of the rain detector are higher in the Mediterranean Sea, where convection is known to be higher \citep{10.1029/2023gl102786}. As northern areas experience scarcer and weaker rain events, the number of rain patches in these regions is lower and leads to a lower confidence in the value of the $F_1$ score. {I}t appears to confirm that the model has better performances on convective events than on stratiform precipitation.

\begin{figure}[]
    \centering
    \includegraphics[width=0.75\linewidth]{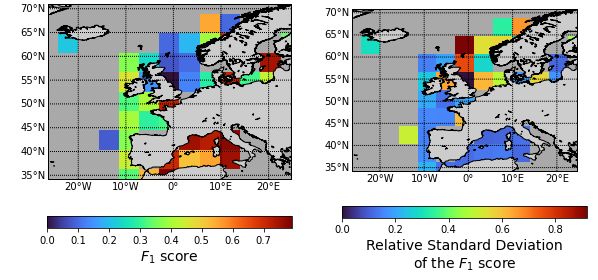}
    \caption{$F_1$ score of the multi-objective model on OPERA data (right) as well as the standard deviation obtained through bootstrapping (left), each with a stride of 4°.}
    \label{fig:eumap}
\end{figure}

\begin{figure*}[]
    \centering
    \includegraphics[width=0.9\linewidth]{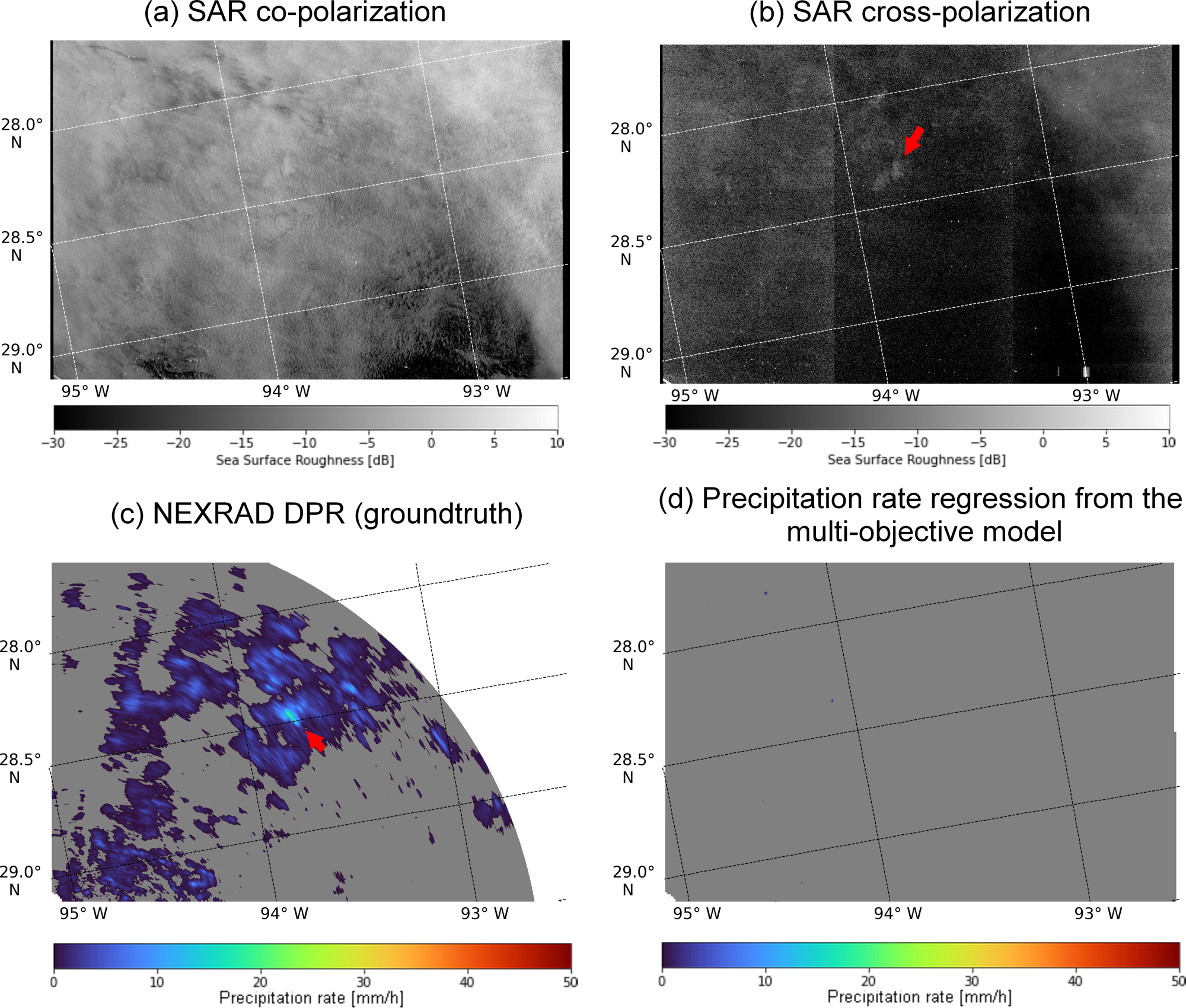}
    
    \caption{Co-observation from Sentinel-1 (acquired on 2020-12-06 at 00:18:19 UTC) and NEXRAD (acquired at 00:19) 2020/12/06 at 00:18:19  UTC.}
    \label{fig:20201206T001819}
\end{figure*}

The underestimate of strat{if}orm rainfall by the regressor can be observed in Figure \ref{fig:20201206T001819}. In this example, no rain signature appears on the co-polarization channel {(a)}. A NEXRAD ground station in the vicinity of the SAR observation observed a rain event with a large spread but low precipitation rate {(c)}. The cross-polarization channel is dominated by noise because of the low wind speed {(b)}. An area, indicated by the red arrow, appears to have slightly higher Sea Surface Roughness but is off by a few kilometers from the maximum DPR provided by NEXRAD. The deep learning model failed to detect this rain event {(d)}. {It is possible that}  the weak impact of the droplets on the ocean surface. Stratiform rains produce smaller and slower droplets. The resulting ring-wave may not be visible on the SAR observation. Smaller droplets in a shallower melting layer may also explain weaker sensitivity to stratiform rain events.

The categorization-based metrics (recall, precision, and $F_1$ score) can be studied by varying the rainfall threshold. In Figure \ref{fig:rr_thresholds_v3}, the precision and the recall are computed for various wind speed range and rainfall threshold. Overall, the precision remains high until the wind exceed 16 m/s, and slightly decreases at higher rainfall thresholds. The impact of the precipitation rate on the recall is higher, with a large increase at higher rainfall. It indicates that the fainer rains are more likely to be underestimated and fall under the decision threshold.

\begin{figure}[]
    \centering
    \includegraphics[width=0.65\linewidth]{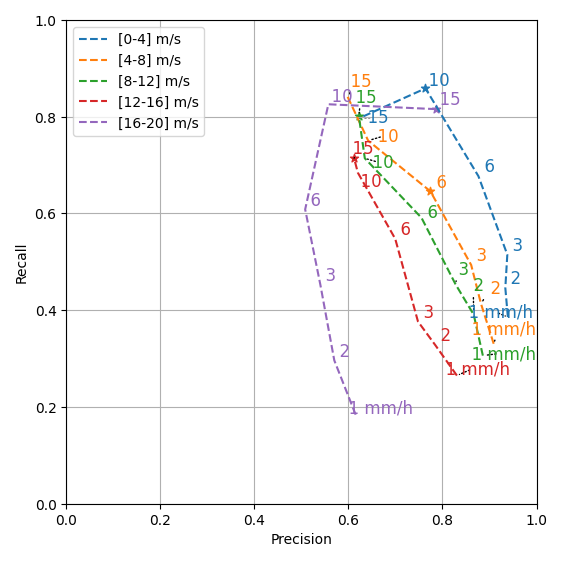}
    \caption{{Performance diagram (recall vs. precision)} of the multi-objective model at different rainfall thresholds {(in mm/h)}. Each curve correspond to a single wind speed interval. {A star indicates, for each wind speed range, the precipitation rate with the highest $F_1$ score.}}
    \label{fig:rr_thresholds_v3}
\end{figure}

\section{Conclusion and Discussion}

A convolutional neural network for the regression of precipitation rate was trained on a large dataset of 29,369 wide-swath observations of Sentinel-1 collocated with NEXRAD. This model is trained in a multi-objective framework to mitigate the discrepancies between both types of sensors. Compared with previous models, several enhancements are proposed to tackle the particularities of the rainfall and wind conditions, as both are long-tailed distributions. In particular, the model is constrained to ensure that the mean and maximum precipitation rates both match the groundtruth, and contains a adversarial loss to provide implicit prior on the groundtruth distribution. The training process is designed to favor scarce meteorological situation and increase the wind speed range available for rain detection.
{However, subsampling at medium wind speed ensure that the higher wind speeds are over-represented with respect to the climatology, which is necessary to extend the model’s capacity to actively learn such conditions. Otherwise, the model would focus the performance improvement over the most frequent wind regime but undermine its relevance at higher winds. As indicated in the Fig. 8, the performance is not hindered at low and medium wind speed, but significantly improved above 8 m/s.}

The model is evaluated both as a rain detector, which enables comparison with {existing rain segmenter}, and as a rain estimator. The rain detector shows a large increase in precision, indicating a lower number of false positives (i.e. overestimates) of the rainfall. The performance remains stable until the wind speed reaches 15 m/s. The version of the {IPF} appears to have a slight impact on the performance, indicating a need to monitor the model's performance in the future. 

The model is able to produce high-resolution maps of the convective rain cells, at a resolution of 200 m/px. It appears to provide valuable information for further studies of the rain cells or in data assimilation frameworks. However, its performance varies geographically, with higher performance obtained in areas with frequent convective activity and at higher precipitation rates. 

\clearpage

\datastatement

Observations from Sentinel-1 can be accessed through the Copernicus Data Space Ecosystem (\url{https://dataspace.copernicus.eu/}). Observations from NEXRAD can be accessed from AWS (\url{https://registry.opendataaws/noaa-nexrad/}). A tool to collocate these sensors is proposed on Github (\url{https://github.com/CIA-Oceanix/Sentinel1-colocalisations}). Samples of collocations and inferences are available on Kaggle (\url{https://www.kaggle.com/datasets/rignak/sentinel1-ocean-precipitation-rate-preview}).

\bibliographystyle{ametsocV6}
\bibliography{references}

\clearpage

\end{document}